\documentclass[11pt,a4paper]{article}
\usepackage[hyperref]{emnlp2020}
\usepackage{times}
\usepackage{latexsym}

\usepackage[normalem]{ulem}

\usepackage{microtype}

\aclfinalcopy 
\def\aclpaperid{18} 

\usepackage{array}
\usepackage{times}
\usepackage{latexsym}
\usepackage{pdflscape}
\usepackage{booktabs}
\usepackage{longtable}
\usepackage{url}
\usepackage{setspace}
\usepackage{hanging}
\usepackage{amsmath}
\usepackage{graphicx}
\usepackage{color, colortbl}
\usepackage{xcolor}
\usepackage[utf8]{inputenc}
\usepackage[ngerman]{babel}

\definecolor{Gray}{gray}{0.84}

\title{Fine-grained linguistic evaluation for state-of-the-art Machine Translation}

\author{Eleftherios Avramidis, Vivien Macketanz, Ursula Strohriegel,\\ \textbf{Aljoscha Burchardt} and \textbf{Sebastian Möller} \\
  German Research Center for Artificial Intelligence (DFKI), Berlin, Germany \\
  {\tt firstname.lastname@dfki.de} \\}

\date{}

\begin{document}
\maketitle
\begin{abstract}
This paper describes a test suite submission providing detailed statistics of linguistic performance for the state-of-the-art German-English systems of the Fifth Conference of Machine Translation (WMT20). 
The analysis covers 107 phenomena organized in 14 categories based on about 5,500 test items, including a manual annotation effort of 45 person hours. 
Two systems (Tohoku and Huoshan) appear to have significantly better test suite accuracy than the others, although the best system of WMT20 is not significantly better than the one from WMT19 in a macro-average. 
Additionally, we identify some linguistic phenomena where all systems suffer (such as idioms, resultative predicates and pluperfect), but we are also able to identify particular weaknesses for individual systems (such as quotation marks, lexical ambiguity and sluicing).
Most of the systems of WMT19 which submitted new versions this year show improvements. 
\end{abstract}

\section{Introduction}

Fine-grained evaluation has recently had increasing interest on several natural language processing (NLP) tasks. 
Focusing on particular issues gives the possibility to analyse the automatic output in ways that cannot be seen by generic metrics. 
This is of particular importance in the era of deep learning, which has led to high performances and differences that are relatively difficult to distinguish. 
Additionally, detailed evaluation can provide indications for the improvement of the systems and the data collection, or allow focusing on phenomena of the long tail that might be of particular interest for certain cases \cite[e.g. social biases;][]{stanovsky-etal-2019-evaluating}.

The most common method for fine-grained or focused evaluation are the \textit{test suites} \cite[also known as \textit{challenge sets} or \textit{benchmarks}; ][]{Guillou2016,isabelle-etal-2017-challenge}.
These are test sets engineered in a particular way, so that they can test the performance of NLP tasks on concrete issues. 

This paper is presenting the use of such a test suite for the evaluation of the 11 German$\rightarrow$English Machine Translation (MT) systems that participated at the Shared Task of the Fifth Conference of Machine Translation (WMT20). 
The evaluation applies the DFKI test suite on German-English, via 5,514 test items which cover 107 linguistically motivated phenomena organized in 14 categories. 
After a reference in related work (Section~\ref{sec:related}), we explain shortly the structure of the test suite (Section~\ref{sec:method}) and present the results (Section~\ref{sec:results}) and the conclusions (Section~\ref{sec:conclusions}).

\section{Related Work}
\label{sec:related}

The use of test suites was introduced along with the early steps of MT in the 1990's \citep{King1990,Way1991,heid1991some}. 
With the emergence of deep learning, recent works re-introduced test suites that focus on the evaluation of particular
linguistic phenomena \cite[e.g.
pronoun translation;][]{Guillou2016} or more generic test suites that aim at
comparing different MT technologies \citep{Isabelle2017,Burchardt2017} and
Quality Estimation methods \cite{avramidis-etal-2018-fine}.
The test suite track of the Conference of Machine Translation has already taken place two years in a row, allowing the presentation of several test suites, focusing on various linguistic phenomena and supporting different language directions.
These include work in grammatical
contrasts \cite{cinkova-bojar:2018:WMT}, discourse
\cite{bojar-EtAl:2018:WMT2}, morphology \cite{burlot-EtAl:2018:WMT}, pronouns \cite{guillou-EtAl:2018:WMT}
and word sense disambiguation \cite{rios-mller-sennrich:2018:WMT}.
When compared to the vast majority of the previous test suites, the one presented here is the only one that performs a systematic evaluation of more than one hundred phenomena on the state-of-the-art systems participating in WMT20.

\section{Method}
\label{sec:method}

The test suite is a test set that has been devised manually with the aim to allow testing the MT output for several linguistic phenomena. 
The entire test suite consists of subsets that test one particular phenomenon each, through several test items.
Each test item of the test suite consists of a source sentence and a set of correct and/or incorrect MT outputs. 
At the evaluation time, the test items are given as input to the MT systems and it is tested on whether the respective MT output consists a correct translation. 
By observing the amount of the test items that are translated correctly, one can calculate the performance of the MT systems regarding the respective phenomenon.

\begin{table}
\centering
\small
\begin{tabular}{ll}
\toprule
\rowcolor{Gray}
Lexical Ambiguity 	& \\
Er las gerne Novellen.	&\\
He liked to read novels.	&fail\\
 He liked to read novellas.	&pass\\
\rowcolor{Gray}
Phrasal verb	& \\
Warum starben die Dinosaurier aus? \\
Why did the dinosaurs die?	&fail\\
Why did the dinosaurs die out?	&pass\\
Why did the dinosaurs become extinct? &pass\\
\rowcolor{Gray}
Ditransitive Perfect	& \\
    Ich habe Tim einen Kuchen gebacken. 	&\\
	I have baked a cake.	&fail\\
	I baked Tim a cake.	&pass\\
\bottomrule
\end{tabular}
\caption{Examples of passing and failing MT outputs}
\label{tab:examples}
\end{table}

The evaluation presented in this paper is based on the DFKI Test Suite for MT on German
to English, which has been presented in \cite{Burchardt2017} and applied
extensively in the WMT shared task of 2018 \cite{Macketanz2018} and 2019 \cite{avramidis-etal-2019-linguistic}.
The current version includes 5,560 test items in order to control 107
phenomena organised in 14 categories.
Some sample test items can be seen in Table~\ref{tab:examples} whereas a more detailed list of test sentences with correct and incorrect translations can be found on GitHub\footnote{\url{https://github.com/DFKI-NLP/TQ_AutoTest}}.


\begin{figure*}
\centering
\includegraphics[width=.9\textwidth]{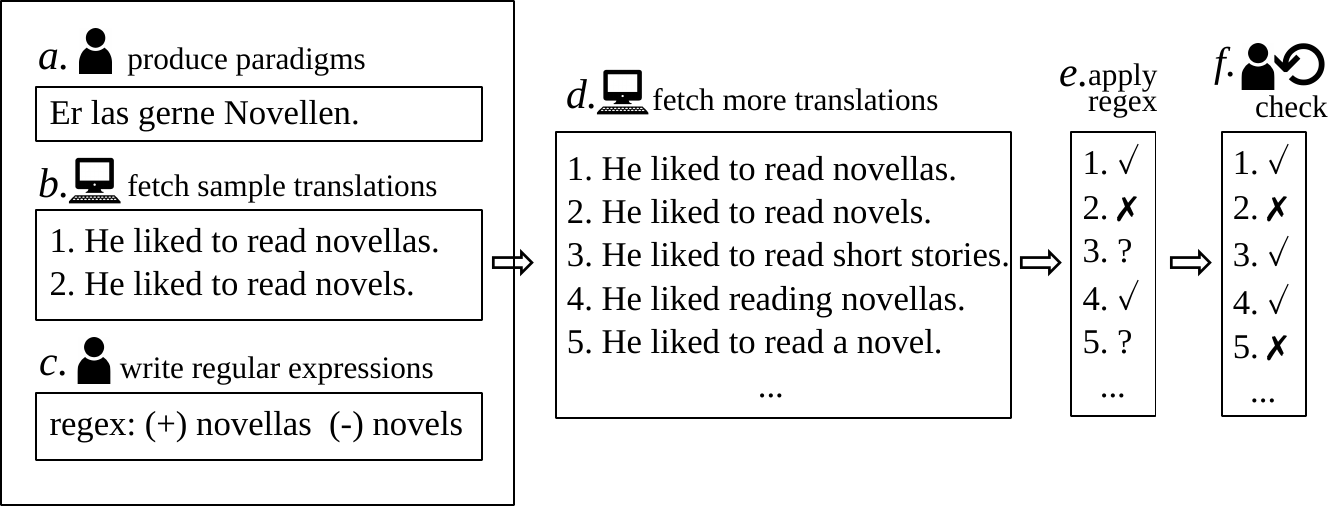}
\caption{Example of the preparation and application of the test suite for one
test sentence}
\label{fig:example}
\end{figure*}


\subsection{Application of the test suite}

The construction of the test suite has been thoroughly explained in the papers from the previous years \cite{avramidis-etal-2018-fine,avramidis-etal-2019-linguistic} and depicted in Figure~\ref{fig:example} (steps \textit{a-c}). 
The test items of the test suite are given as input to the MT systems (step \textit{d}). 
Their MT outputs are tested using a set of rules (regular expressions or fixed strings), each rule specific for a phenomenon, that defines whether the translations are correct with respect to the tested phenomenon (step \textit{e}). 
When the automatic application of the rules cannot lead to a clear decision on whether the translation is correct or not, the test item is left with a warning. 
The warnings are consequently resolved by human annotators with linguistic knowledge, who inspect the MT output, provide a clear judgment and also augment the set of the rules to cover similar cases in the future (step \textit{e}). 

For every system we calculate the phenomenon-specific translation accuracy as the the number of the test sentences for the phenomenon which were translated properly, divided by the number of all test sentences for this phenomenon: 

\begin{equation*} 
 \textrm{accuracy} =  \frac{\textrm{correct\;translations}}{\textrm{sum\;of\;test\;items}} 
\end{equation*}

Each phenomenon is covered by at least 20 test items\footnote{with the exception of 7 phenomena which have 9-19 items}, whereas the same test items are given to multiple systems to achieve comparisons among them.
In order to achieve a fair comparison among the systems, only the test items that do not contain any warnings for any of the systems are included in the calculation.

In order to define which systems have the best performance for a particular phenomenon, all systems are compared with the system with the highest accuracy.
When comparing the highest scoring system with the rest, the significance of the comparison is confirmed with a one-tailed Z-test with $\alpha = 0.95$.
The systems whose difference with the best system is not significant are considered to be in the first performance cluster and indicated with boldface in the tables. 

\subsection{Experiment setup}

In the evaluation presented in the paper, MT outputs are obtained from the 11
systems that are part of the \textit{news translation task} of the Fifth
Conference on Machine Translation (WMT20). 
These are 7 systems submitted by the shared task participants plus 4 online commercial systems whose output has been obtained by the workshop organizers and therefore have been anonymized. 
Unfortunately, contrary to previous years, very few system descriptions were provided by the time this paper was written and it is therefore not possible to associate linguistic performance with system types and settings. 

As explained earlier, the application of the test suite on the output of the 11 systems left about 10\% of unresolved warnings which needed to be manually edited. 
A human annotator with linguistic background devoted about 45 working hours in order to resolve 99\% of them (resulting into 5,514 valid out of 5,560 total items).

\section{Results}
\label{sec:results}

The accuracy of each system per linguistic category is briefly shown in Table~\ref{tab:categories} whereas the detailed statistics depicting the accuracy for every linguistic phenomenon, grouped in the respective linguistic categories are shown in Table~\ref{tab:phenomena}. 
Since every category and every phenomenon have a different amount of test items, the average scores, shown in the last rows of the tables, are computed in three different ways: 
The first aggregates the contributions of all test items to compute the average percentages (micro-average), the second (Table~\ref{tab:categories}) computes the percentages independently for each category and then takes the average (hence treating all categories equally; \textit{category macro-average}) and the third (Table~\ref{tab:phenomena}) computes the percentages independently for each phenomenon and then takes the average (hence treating all phenomena equally; \textit{phenomenon macro-average}). 
The significantly best systems for every category or phenomenon are bold-faced. 

Very high scores do not necessarily mean that the MT of the respective grammatical phenomenon has been solved, but rather that the current test items of the test suite (which was engineered with the emergence of the first neural MT systems in 2017) are unable to expose difficulties of the systems. 
The artificial nature of the test suite and the variable number of test items per category and phenomenon should also be taken in consideration when doing comparisons between categories and phenomena.

\subsection{Comparison between systems}

Two systems are standing out for their overall performance. 
\textbf{Tohoku}-AIP-NTT \cite{Kiyono2020} achieves the best category macro-averaged accuracy of 88.1\%, whereas together with \textbf{Huoshan} they share the first position based on their micro-averaged accuracy (85.3-85.4\%). 
The systems UEdin \cite{Germann2020}, Online-B, Online-G and Online-A are next. 

Tohoku and Huoshan are the best performing systems for all linguistic categories, whereas UEdin is losing in one category and Online-A is losing in two categories.

Two systems, WMTBiomedBaseline and ZLabs show very low performances and are assumed to be non state-of-the-art systems.
We will therefore exclude these two systems from the discussion and conclusions for phenomena and categories. 
Whereas no description was available for ZLab, the lower performance of the WMTBiomedBaseline can be attributed to the fact that it was trained with only 56\% of the parallel training data used by Tohoku and no synthetic data.
Among the rest of the systems, the worst performing one is Online-Z, achieving the lowest accuracy (74\% on both micro- and macro-average), being on par with the best systems on only 6 categories. 

BLEU scores \cite{papineni-etal-2002-bleu} on the official test-set are also calculated for further comparison. The order of the systems based on BLEU seems to correlate with the order given by the category macro-average with the exception of one system \cite[OPPO; ][]{Shi2020}. 

\subsection{Linguistic categories}

The average accuracy regarding the linguistic categories ranges in relatively high numbers, between 68.9\% and 97.3\%.
The categories with the highest accuracy in average are the \textbf{negation} (97.3\%), the \textbf{composition} (85.3\%), the \textbf{subordination} (85.3\%) and the \textbf{named entinties and terminology} (82\%). 
The ones with the lowest accuracy are the \textbf{multi-word expressions} (MWE), the \textbf{ambiguity}, the \textbf{false friends} and the \textbf{verb valency} (68.9-71.5\%).

When one tries to identify weaknesses of particular systems, OPPO is suffering mostly concerning function words and long distance dependencies (LDD) / interrogatives. 
Online-Z and PROMT have issues with ambiguity and several systems have issues with punctuation.
Some of these issues are discussed in a more fine-grained level below. 

The comparison of the state-of-the-art systems with the low-resource WMTBiomedBaseline indicates that some categories, such as ambiguity and composition, are particularly sensitive to low resources, as their accuracy is proportionally lower than other categories if compared to the respective category accuracies of the state-of-the-art systems.

\begin{table}
\centering
\small
\begin{tabular}{ll}
\toprule
\rowcolor{Gray}
Verb Valency 	& \\
Ich erinnere mich seiner.	&\\
I remember his.	&fail\\
I remember him.	&pass\\
\rowcolor{Gray}
False Friends	& \\
Er las gerne Novellen. \\
He liked to read novels. 	&fail\\
He liked to read novellas.	&pass\\
\bottomrule
\end{tabular}
\caption{Examples of linguistic categories with lower accuracy with passing and failing MT outputs}
\label{tab:examplescat}
\end{table}

Table \ref{tab:examplescat} contains examples from the two low accuracy categories \textit{verb valency} and \textit{false friends}. \textit{Verb valency} refers to the arguments that are being controlled by the predicate. Certain verbs require a specific grammatical case. In our example, the German verb \textit{sich erinnern} (to remember) requires a genitive object, in this case \textit{seiner}. \textit{Seiner}, however, can also mean \textit{his} as in the possessive pronoun, which explains the mistranslation of \textit{I remember his}. 

\textit{False friends} are words in different languages that look similar and are therefore often mistaken for being translations of one another, even though their meanings differ. The German noun \textit{Novelle} does not translate to \textit{novel}, but to \textit{novella} or \textit{short story}. While you would expect a human to make these kind of translation errors, it is surprising to see that also MT systems are prone to mistranslating false friends.

\subsection{Linguistic phenomena}

The accuracy regarding individual linguistic phenomena has a wide range, between very low scores (15\%) and full success (100\%). 
The phenomena which all systems had difficulty to handle were the \textbf{idioms} and the \textbf{resultative predicates}, with most systems scoring only 20\% and 26\% respectively. 
However, the overall performance on these phenomena has improved: last year only 3 systems could achieve this performance, with the majority of the systems having 5-10\% less accuracy.
\textbf{Modal pluperfect} is also ranging very low, scoring between 2.2\% and 50.6\% and similar is the case for its negated version.
Other moods of the pluperfect make it particularly difficult for some systems, e.g. PROMT and Online-Z suffer in translating the \textbf{ditransitive} and \textbf{intransitive pluperfect}.

When trying to find the cases that consist a weakness for particular systems, Online-Z indicates one of the lowest scores in punctuation, which appears to derive from the fact that the system removes all \textbf{quotation marks}. 
A similar issue is observed with Online-G and OPPO which could correctly convey almost half of the quotation marks, whereas another two systems have some way to go.
Interestingly enough, despite strongly depending on preprocessing, quotation marks are a common issue, since similar cases have been noted in previous years. 
In other phenomena, Online-Z has a very low accuracy for \textbf{sluicing} whereas PROMT is relatively weak concerning \textbf{lexical ambiguity}.

\begin{table}
\centering
\small
\begin{tabular}{ll}
\toprule
\rowcolor{Gray}
Resultative Predicate 	& \\
Sie trinkt die Tasse leer.	&\\
She drinks the cup empty.	&fail\\
\textit{She empties the cup.}	&\textit{pass}\\
\textit{She is drinking the whole cup.}  &\textit{pass} \\
\rowcolor{Gray}
Intransitive Pluperfect	& \\
Sie hatten geschlafen. \\
They were sleeping. 	&fail\\
They had slept.	&pass\\
\rowcolor{Gray}
Sluicing	& \\
John mag die Nudeln nicht, aber er weiß \\nicht, warum. \\
John doesn't like the noodles but he doesn't \\know why. 	&fail\\
John doesn't like the noodles, but he doesn't \\know why.	&pass\\
\rowcolor{Gray}
Lexical Ambiguity	& \\
Das Gericht gestern Abend war lecker. \\
The court last night was delicious. 	&fail\\
The dish last night was delicious.	&pass\\
\bottomrule
\end{tabular}
\caption{Examples of linguistic phenomena with low accuracy with passing and failing MT outputs}
\label{tab:examplesphen}
\end{table}

Table \ref{tab:examplesphen} contains further translation examples from linguistic phenomena with low accuracy.
A \textit{resultative predicate} is a construction that consists of a verb and an adjective in which the verb describes an action and the adjective describes the result of that action. In many cases, resultative predicates lead to translation errors as they do not exist in English and a literal translation leads to an ungrammatical translation, as can be seen in the example. Since none of the systems could produce a correct output for this sentence, we have provided two possible correct translations here as examples. 

\textit{Intransitive verbs} do not require further objects (as opposed to transitive or ditransitive verbs). \textit{Pluperfect} is a tense which is used in German to describe completed actions that have taken place in the past. It should be translated to English in pluperfect as well. In the example, the incorrect translation contains past progressive \textit{were sleeping} instead of the correct \textit{had slept}.

\textit{Sluicing} is a type of ellipsis that can occur in direct and indirect interrogative clauses. A wh-word precedes the part of the sentence that contains the ellipsis. In our example, all constituents following the wh-word are elided: \textit{John mag die Nudeln nicht, aber er wei{\ss}  nicht, warum  \sout{er die Nudeln nicht mag}.} Sluicing exists in both German and in English: \textit{John doesn't like the noodles, but he doesn't know why  \sout{he doesn't like the noodles}.} One difference between the German and the English sluicing sentence is that in German there are two commas, while in English  there is only one. Since this phenomenon concerns the complete sentence, punctuation should be correct when translating a sentence containing sluicing. In our example, the missing comma leads to fail. 

The fourth example in the Table contains the lexical ambiguity \textit{Gericht}. \textit{Gericht} can either mean \textit{court} or \textit{dish} but  the context provided in the sentence (\textit{war lecker}, English: \textit{was delicious}) serves as disambiguation so that only a translation referring to \textit{dish} (or to food in some way) can be a pass. Any translation referring to \textit{court/courthouse/tribunal} or the like is a fail. 

\subsection{Comparison with previous years}

\begin{table}[]
    \centering
    \begin{tabular}{ccc}
    \toprule
         WMT & categ. macro- & micro- \\
    \midrule
         2018 & 81.0 & \textbf{84.1} \\
         2019 & \textbf{87.4} & 83.0 \\
         2020 & \textbf{88.0} & \textbf{85.1} \\
    \bottomrule
    \end{tabular}
    \caption{The accuracy (\%) of the best system of each year as measured over 5,555 test items that were common over the last years. The scores that are significantly higher in each column are boldfaced}
    \label{tab:bestsystems}
\end{table}

One can notice some improvements on the overall performance of the best system, as compared with the previous two years. 
As seen in Table~\ref{tab:bestsystems} from 2018 to 2019 there was a 6.4\% improvement on the macro-averaged accuracy but there was no significant improvement from 2019 to 2020.
The best system of 2019 was not submitted in 2020 and this is unfortunate, as it performed better than this year's best system in four categories (mostly regarding ambiguity; Table~\ref{tab:compareyears}).
When considering micro-averaged accuracy, there is significant improvement since last year (2.1\%), but the best system of 2018 is competing with the one of this year, due to its high performance regarding verb tense/mood.

One can also consider the improvement of individual systems submitted to WMT from one year to another, starting from 2018. 
This year, only 5 of the 2019 systems submitted their new version and the yearly difference of their test suite accuracy can be seen in Table~\ref{tab:compareyears}. 
The accuracy is measured over the test items that are common over all three years (or at least the last two).

All systems indicate considerable improvements on the macro-average since the previous year, ranging between 2,4\% and 8,5\%.
Online-G had a  major improvement for a second year in a row (21.6\% in total), whereas it is the only system that achieved such an improvement without having an accuracy drop for any of the linguistic categories, whereas it improved 6 categories for more than 10\%. 
PROMT had improvement in all categories apart from verb tense/aspect/mood, UEdin deteriorated in verb tense/aspect/mood, whereas Online-B deteriorated in composition, named entity/terminology and punctuation. 

The linguistic categories that improved mostly in average are the \textbf{long distance dependencies / interrogatives},  the \textbf{verb valency}, the \textbf{ambiguity} and the \textbf{punctuation}. 

\section{Conclusions and further work}
\label{sec:conclusions}

In this paper we present the results of the application of the DFKI test suite in the output of the state-of-the-art MT systems participating in the Shared Task of the Fifth Conference of Machine Translation (WMT20).
Based on about 5,500 test items, we present detailed accuracies regarding 107 phenomena organized in 14 categories. 
Additionally, the evolution of systems submitted also in previous years is observed. 

The best system of this year is not significantly better than the one from 2019 in a macro-average, but one can see significant improvement from two years ago. 
The systems that seem to have the best accuracies are Tohoku and Huoshan.
The phenomena that most systems face difficulties are again this year the idioms, the resultative predicates and some moods of the pluperfect, whereas some systems still have issues with quotation marks and lexical ambiguity.

As discussed previously, the high accuracies achieved for particular phenomena or categories raise questions on whether these phenomena are getting solved, or whether the test suite (which was originally built to challenge the systems from 2017) should raise the difficulty by including more test items. 

In further work, we would like to be able to associate the performance on specific phenomena with  decisions related to decisions during the development of the systems, once there is enough information about this process for all systems. 

\section*{Acknowledgments}

This research was supported by the German Research Foundation through the project TextQ and by the German Federal Ministry of Education through the project SocialWear.

\bibliography{final}
\bibliographystyle{acl_natbib}

\onecolumn
\appendix
\begin{landscape}

{\footnotesize
    
\begin{longtable}[c]{lrrrrrrrrrrrrr}
\toprule
category & items & Tohoku & Huoshan & UEdin & Onl-B & Onl-G & Onl-A & PROMT & OPPO & Onl-Z & ZLabs & WMTBi & avg \\
\midrule
Ambiguity                     &   81 &  \textbf{82.7} &  \textbf{77.8} &  \textbf{72.8} &  \textbf{79.0} &  \textbf{84.0} &  \textbf{76.5} &          64.2  &  \textbf{82.7} &          67.9  &          45.7  &          30.9  & 69.5 \\
Composition                   &   49 &  \textbf{98.0} &  \textbf{98.0} &  \textbf{93.9} &  \textbf{93.9} &  \textbf{95.9} &  \textbf{93.9} &  \textbf{89.8} &  \textbf{95.9} &          85.7  &          49.0  &          44.9  & 85.3 \\
Coordination \& ellipsis      &   78 &  \textbf{89.7} &  \textbf{91.0} &  \textbf{89.7} &  \textbf{91.0} &  \textbf{85.9} &  \textbf{87.2} &  \textbf{87.2} &  \textbf{87.2} &          60.3  &          52.6  &          44.9  & 78.8 \\
False friends                 &   36 &  \textbf{72.2} &  \textbf{80.6} &  \textbf{72.2} &  \textbf{80.6} &  \textbf{77.8} &  \textbf{69.4} &  \textbf{72.2} &          66.7  &  \textbf{86.1} &          52.8  &          50.0  & 71.0 \\
Function word                 &   72 &  \textbf{86.1} &  \textbf{80.6} &  \textbf{86.1} &  \textbf{90.3} &  \textbf{90.3} &  \textbf{83.3} &  \textbf{88.9} &          55.6  &  \textbf{88.9} &          41.7  &          43.1  & 75.9 \\
LDD \& interrogatives         &  174 &  \textbf{89.1} &  \textbf{86.2} &  \textbf{85.1} &  \textbf{83.3} &  \textbf{86.8} &          77.6  &          81.0  &          58.6  &          72.4  &          48.3  &          58.6  & 75.2 \\
MWE                           &   80 &  \textbf{80.0} &  \textbf{75.0} &  \textbf{71.3} &  \textbf{77.5} &  \textbf{77.5} &  \textbf{71.3} &  \textbf{70.0} &  \textbf{78.8} &  \textbf{73.8} &          45.0  &          37.5  & 68.9 \\
Named entitiy \& terminology  &   89 &  \textbf{92.1} &  \textbf{84.3} &  \textbf{87.6} &          82.0  &          82.0  &  \textbf{88.8} &  \textbf{87.6} &  \textbf{85.4} &          68.5  &          70.8  &          73.0  & 82.0 \\
Negation                      &   20 & \textbf{100.0} & \textbf{100.0} & \textbf{100.0} & \textbf{100.0} & \textbf{100.0} &  \textbf{95.0} & \textbf{100.0} & \textbf{100.0} &  \textbf{95.0} &          80.0  & \textbf{100.0} & 97.3 \\
Non-verbal agreement          &   61 &  \textbf{91.8} &  \textbf{88.5} &  \textbf{88.5} &  \textbf{86.9} &  \textbf{90.2} &  \textbf{83.6} &  \textbf{82.0} &  \textbf{88.5} &  \textbf{85.2} &          54.1  &          57.4  & 81.5 \\
Punctuation                   &   60 &  \textbf{96.7} &  \textbf{98.3} &  \textbf{98.3} &          71.7  &          61.7  & \textbf{100.0} &  \textbf{98.3} &          70.0  &          28.3  &          68.3  &          55.0  & 77.0 \\
Subordination                 &  180 &  \textbf{90.6} &  \textbf{88.3} &  \textbf{91.1} &  \textbf{91.1} &  \textbf{92.2} &  \textbf{88.9} &  \textbf{90.0} &  \textbf{90.6} &  \textbf{87.8} &          65.0  &          62.2  & 85.3 \\
Verb tense/aspect/mood        & 4447 &  \textbf{84.6} &  \textbf{85.3} &          80.3  &          75.9  &          79.6  &          77.5  &          75.1  &          79.3  &          73.6  &          50.5  &          52.1  & 74.0 \\
Verb valency                  &   87 &  \textbf{79.3} &  \textbf{81.6} &  \textbf{77.0} &  \textbf{81.6} &  \textbf{77.0} &  \textbf{77.0} &  \textbf{71.3} &  \textbf{80.5} &          64.4  &          44.8  &          51.7  & 71.5 \\
\midrule
micro-average                 & 5514 &  \textbf{85.3} &  \textbf{85.4} &          81.2  &          77.7  &          80.6  &          78.7  &          76.5  &          79.1  &          73.6  &          51.3  &          52.4  & 74.7 \\
macro-average                 & 5514 &  \textbf{88.1} &          86.8  &          85.3  &          84.6  &          84.3  &          83.6  &          82.7  &          80.0  &          74.1  &          54.9  &          54.4  & 78.1 \\
BLEU & & 43.8 & 43.5 & 42.3 & 41.9 & 41.4 & 40.4 & 39.6 & 43.2 & 35.4 & 31.5 & 32.1 & 35.4 \\
\bottomrule

\caption{Accuracies (\%) of successful translations for 11 systems and 14
      categories. Boldface indicates the significantly best performing systems in each row}
\label{tab:categories}
\end{longtable}

\vspace{-15px}

\begin{longtable}{l | rr | rr | rr | r | rr | rr }
    \toprule
	& \multicolumn{2}{c|}{Onl-A}	&  \multicolumn{2}{c|}{Onl-B}	&  \multicolumn{2}{c|}{Onl-G} & PROMT	&  \multicolumn{2}{c|}{UEdin}  &  \multicolumn{2}{c}{best} \\
category	&2019	&2020	&2019	&2020	&2019	&2020	&  \multicolumn{1}{c|}{2020}	&2019 &2020 &2019 &2020 \\
\midrule

Ambiguity	&+2.6	&+7.7	&+1.3	&+2.6	&+2.6	&+11.5	&+16.7	&+11.6	&+14.1 & +12.4 &	\textcolor{red}{-9.9}\\
Composition	&+10.4	&+2.1	&	&\textcolor{red}{-4.1}	&+12.5	&+12.5	&+10.4	&+8.3	&+8.4 & +12.3 & \\
Coordination \& ellipsis	&	&	&	&+2.2	&+8.7	&+19.6	&+6.6	&	&+4.4 & 5.1 & \textcolor{red}{-1.3} \\
False friends	&	&\textcolor{red}{-2.8}	&+2.8	&+2.8	&	&+5.6	&	&+13.9	&+5.5 &+8.3 &\textcolor{red}{-2.8}\\
Function word	&+1.6	&\textcolor{red}{-1.6}	&	&+10.9	&+42.2	&+1.6	&+6.2	&+7.8	& +2.7 & \\
LDD \& interrogatives	&\textcolor{red}{-2.2}	&+8.7	&	&+6.6	&+14.2	&+18.4	&+8.7	&	&+15.2 & +1.8 & +3.4\\
MWE	&+1.3	&+5.4	&	&+6.6	&+5.3	&+9.3	&+14.6	&+5.3	&+10.7 & +10.0 & +1.2 \\
Named entitiy \& terminology	&	&+3.1	&	&\textcolor{red}{-3.2}	&\textcolor{red}{-1.5}	&+7.8	&+15.6	&+7.8	&+3.1 & \textcolor{red}{-2.3} &	+9.0 \\
Negation	&	&\textcolor{red}{-5.0}	&	&+5.0	&+4	&	&+10.0	&\textcolor{red}{-10.0}	&+10.0 & & \\
Non-verbal agreement	&+6.9	&	&	&	&+22.4	&+10.4	&+10.4	&+10.4	&+6.9 & +9.8 &  \\
Punctuation	&\textcolor{red}{-21.9}	&+25.5	&	&\textcolor{red}{-3.6}	&\textcolor{red}{-7.3}	&+3.7	&+16.4	&+21.8	&+9.1 & +30.0 &	+3.4 \\
Subordination	&\textcolor{red}{-11.3}	&+10.3	&+1.0	&+7.2	&+6.2	&+5.1	&+4.1	&\textcolor{red}{-2.1}	&+7.2 & +3.8 & +1.2\\
Verb tense/aspect/mood	&+11.9	&\textcolor{red}{-6.3}	&+0.2	&+2.0	&+19.8	&+13.9	&\textcolor{red}{-0.2}	&+5.6	&\textcolor{red}{-1.1} & 0.8 & 5.8 \\
Verb valency	&+1.5	&+9.0	&	&+13.5	&+11.9	&+6.0	&+11.9	&+3.0	&+13.4 & 9.2 & \textcolor{red}{-1.2} \\
\midrule
micro-avg	&+9.9	&\textcolor{red}{-4.4}	&+0.2	&+2.4	&+18.5	&+13.1	&+1.5	&+5.6	&+0.6 & +2.0 & +4.8
\\
macro-avg	&+0.1	&+4.0	&+0.4	&+3.5	&+12.7	&+8.9	&+9.3	&+6.0	&+7.6 &  +7.4 & +0.6\\
\bottomrule
  
    \caption{Difference of the test suite accuracy from one year to the next one per category, for the systems participating in the shared tasks WMT18-20, measured over the test items that are common over all these years.}
    \label{tab:compareyears}
\end{longtable}

    \begin{longtable}[c]{lrrrrrrrrrrrrr}
\toprule
phenomenon & items & Tohoku & Huoshan & UEdin & Onl-B & Onl-G & Onl-A & PROMT & OPPO & Onl-Z & ZLabs & WMTBi & avg \\
\midrule
\endfirsthead
\toprule
phenomenon & items & Tohoku & Huoshan & UEdin & Onl-B & Onl-G & Onl-A & PROMT & OPPO & Onl-Z & ZLabs & WMTBi & avg \\
\midrule
\endhead
\rowcolor{Gray}
Ambiguity                     &   81 &  \textbf{82.7} &  \textbf{77.8} &  \textbf{72.8} &  \textbf{79.0} &  \textbf{84.0} &  \textbf{76.5} &          64.2  &  \textbf{82.7} &          67.9  &          45.7  &          30.9  & 69.5 \\
Lexical ambiguity             &   63 &  \textbf{85.7} &  \textbf{79.4} &  \textbf{77.8} &  \textbf{77.8} &  \textbf{87.3} &  \textbf{79.4} &          65.1  &  \textbf{85.7} &          68.3  &          49.2  &          36.5  & 72.0 \\
Structural ambiguity          &   18 &  \textbf{72.2} &  \textbf{72.2} &  \textbf{55.6} &  \textbf{83.3} &  \textbf{72.2} &  \textbf{66.7} &  \textbf{61.1} &  \textbf{72.2} &  \textbf{66.7} &          33.3  &          11.1  & 60.6 \\
\rowcolor{Gray}
Composition                   &   49 &  \textbf{98.0} &  \textbf{98.0} &  \textbf{93.9} &  \textbf{93.9} &  \textbf{95.9} &  \textbf{93.9} &  \textbf{89.8} &  \textbf{95.9} &          85.7  &          49.0  &          44.9  & 85.3 \\
Compound                      &   29 &  \textbf{96.6} &  \textbf{96.6} &  \textbf{89.7} &  \textbf{93.1} &  \textbf{93.1} &  \textbf{89.7} &  \textbf{86.2} &  \textbf{96.6} &  \textbf{82.8} &          51.7  &          48.3  & 84.0 \\
Phrasal verb                  &   20 & \textbf{100.0} & \textbf{100.0} & \textbf{100.0} &  \textbf{95.0} & \textbf{100.0} & \textbf{100.0} &  \textbf{95.0} &  \textbf{95.0} &  \textbf{90.0} &          45.0  &          40.0  & 87.3 \\
\rowcolor{Gray}
Coordination \& ellipsis      &   78 &  \textbf{89.7} &  \textbf{91.0} &  \textbf{89.7} &  \textbf{91.0} &  \textbf{85.9} &  \textbf{87.2} &  \textbf{87.2} &  \textbf{87.2} &          60.3  &          52.6  &          44.9  & 78.8 \\
Gapping                       &   20 &  \textbf{95.0} &  \textbf{95.0} &  \textbf{95.0} &  \textbf{95.0} &  \textbf{95.0} &  \textbf{80.0} &  \textbf{95.0} &  \textbf{90.0} &  \textbf{75.0} &          70.0  &          40.0  & 84.1 \\
Right node raising            &   20 &  \textbf{80.0} &  \textbf{85.0} &  \textbf{85.0} &  \textbf{80.0} &  \textbf{65.0} &  \textbf{85.0} &  \textbf{80.0} &  \textbf{85.0} &  \textbf{85.0} &          40.0  &          15.0  & 71.4 \\
Sluicing                      &   18 & \textbf{100.0} & \textbf{100.0} & \textbf{100.0} & \textbf{100.0} & \textbf{100.0} & \textbf{100.0} & \textbf{100.0} & \textbf{100.0} &          11.1  &          72.2  & \textbf{100.0} & 89.4 \\
Stripping                     &   20 &  \textbf{85.0} &  \textbf{85.0} &  \textbf{80.0} &  \textbf{90.0} &  \textbf{85.0} &  \textbf{85.0} &  \textbf{75.0} &  \textbf{75.0} &  \textbf{65.0} &          30.0  &          30.0  & 71.4 \\
\rowcolor{Gray}
False friends                 &   36 &  \textbf{72.2} &  \textbf{80.6} &  \textbf{72.2} &  \textbf{80.6} &  \textbf{77.8} &  \textbf{69.4} &  \textbf{72.2} &          66.7  &  \textbf{86.1} &          52.8  &          50.0  & 71.0 \\
\rowcolor{Gray}
Function word                 &   72 &  \textbf{86.1} &  \textbf{80.6} &  \textbf{86.1} &  \textbf{90.3} &  \textbf{90.3} &  \textbf{83.3} &  \textbf{88.9} &          55.6  &  \textbf{88.9} &          41.7  &          43.1  & 75.9 \\
Focus particle                &   24 &  \textbf{91.7} &  \textbf{91.7} &  \textbf{95.8} &  \textbf{95.8} & \textbf{100.0} &  \textbf{91.7} &  \textbf{95.8} &  \textbf{91.7} &  \textbf{95.8} &          75.0  &          83.3  & 91.7 \\
Modal particle                &   29 &  \textbf{75.9} &  \textbf{72.4} &  \textbf{72.4} &  \textbf{79.3} &  \textbf{75.9} &  \textbf{72.4} &  \textbf{79.3} &  \textbf{62.1} &  \textbf{79.3} &          41.4  &          37.9  & 68.0 \\
Question tag                  &   19 &  \textbf{94.7} &          78.9  &  \textbf{94.7} & \textbf{100.0} & \textbf{100.0} &  \textbf{89.5} &  \textbf{94.7} &           0.0  &  \textbf{94.7} &           0.0  &           0.0  & 67.9 \\
\rowcolor{Gray}
LDD \& interrogatives         &  174 &  \textbf{89.1} &  \textbf{86.2} &  \textbf{85.1} &  \textbf{83.3} &  \textbf{86.8} &          77.6  &          81.0  &          58.6  &          72.4  &          48.3  &          58.6  & 75.2 \\
Extended adjective construction &   20 &  \textbf{85.0} &  \textbf{80.0} &  \textbf{75.0} &  \textbf{80.0} &  \textbf{80.0} &  \textbf{70.0} &  \textbf{75.0} &  \textbf{80.0} &          55.0  &          55.0  &          50.0  & 71.4 \\
Extraposition                 &   20 &          75.0  &          70.0  &          75.0  &          55.0  &          65.0  &          65.0  &          75.0  &          65.0  &          65.0  &          40.0  &          40.0  & 62.7 \\
Multiple connectors           &   20 &  \textbf{90.0} &  \textbf{90.0} &  \textbf{95.0} &  \textbf{75.0} &  \textbf{95.0} &  \textbf{75.0} &          70.0  &  \textbf{80.0} &          65.0  &  \textbf{85.0} &  \textbf{85.0} & 82.3 \\
Pied-piping                   &   20 &  \textbf{90.0} &  \textbf{85.0} &  \textbf{90.0} &  \textbf{90.0} &  \textbf{85.0} &  \textbf{80.0} &  \textbf{90.0} &  \textbf{95.0} &          65.0  &          30.0  &          55.0  & 77.7 \\
Polar question                &   20 & \textbf{100.0} & \textbf{100.0} & \textbf{100.0} & \textbf{100.0} & \textbf{100.0} & \textbf{100.0} &  \textbf{95.0} &          10.0  &          85.0  &          55.0  &          85.0  & 84.5 \\
Scrambling                    &   20 &  \textbf{90.0} &  \textbf{75.0} &  \textbf{70.0} &  \textbf{85.0} &  \textbf{85.0} &          50.0  &  \textbf{65.0} &  \textbf{85.0} &          60.0  &          20.0  &          35.0  & 65.5 \\
Topicalization                &   19 &  \textbf{89.5} &  \textbf{78.9} &  \textbf{78.9} &  \textbf{78.9} &  \textbf{78.9} &  \textbf{78.9} &  \textbf{68.4} &  \textbf{84.2} &  \textbf{73.7} &          36.8  &          36.8  & 71.3 \\
Wh-movement                   &   35 &          91.4  & \textbf{100.0} &          91.4  &  \textbf{94.3} &  \textbf{97.1} &          91.4  &  \textbf{97.1} &           8.6  &  \textbf{94.3} &          57.1  &          71.4  & 81.3 \\
\rowcolor{Gray}
MWE                           &   80 &  \textbf{80.0} &  \textbf{75.0} &  \textbf{71.3} &  \textbf{77.5} &  \textbf{77.5} &  \textbf{71.3} &  \textbf{70.0} &  \textbf{78.8} &  \textbf{73.8} &          45.0  &          37.5  & 68.9 \\
Collocation                   &   20 & \textbf{100.0} &  \textbf{90.0} &          80.0  &  \textbf{95.0} &  \textbf{95.0} &          75.0  &          80.0  &  \textbf{95.0} &  \textbf{90.0} &          35.0  &          20.0  & 77.7 \\
Idiom                         &   20 &  \textbf{25.0} &  \textbf{20.0} &  \textbf{15.0} &  \textbf{20.0} &  \textbf{20.0} &  \textbf{15.0} &   \textbf{5.0} &  \textbf{20.0} &  \textbf{20.0} &           0.0  &   \textbf{5.0} & 15.0 \\
Prepositional MWE             &   20 &  \textbf{95.0} &  \textbf{95.0} &  \textbf{95.0} &  \textbf{95.0} &  \textbf{95.0} &  \textbf{95.0} &  \textbf{95.0} & \textbf{100.0} &  \textbf{90.0} &          70.0  &          60.0  & 89.5 \\
Verbal MWE                    &   20 & \textbf{100.0} &  \textbf{95.0} &  \textbf{95.0} & \textbf{100.0} & \textbf{100.0} & \textbf{100.0} & \textbf{100.0} & \textbf{100.0} &  \textbf{95.0} &          75.0  &          65.0  & 93.2 \\
\rowcolor{Gray}
Named entity \& terminology  &   89 &  \textbf{92.1} &  \textbf{84.3} &  \textbf{87.6} &          82.0  &          82.0  &  \textbf{88.8} &  \textbf{87.6} &  \textbf{85.4} &          68.5  &          70.8  &          73.0  & 82.0 \\
Date                          &   20 & \textbf{100.0} & \textbf{100.0} & \textbf{100.0} &          85.0  & \textbf{100.0} & \textbf{100.0} & \textbf{100.0} &  \textbf{95.0} &          60.0  &          75.0  & \textbf{100.0} & 92.3 \\
Domainspecific term           &   20 &          75.0  &          60.0  &          60.0  &          65.0  &          60.0  &          70.0  &          65.0  &          70.0  &          50.0  &          50.0  &          40.0  & 60.5 \\
Location                      &   20 &          95.0  &          95.0  &          95.0  &          95.0  &          90.0  &          95.0  &          90.0  &          95.0  &          90.0  &          85.0  &          90.0  & 92.3 \\
Measuring unit                &   20 & \textbf{100.0} &  \textbf{90.0} & \textbf{100.0} &  \textbf{90.0} &  \textbf{90.0} & \textbf{100.0} & \textbf{100.0} &  \textbf{90.0} &          75.0  &          85.0  &          80.0  & 90.9 \\
Proper name                   &    9 &  \textbf{88.9} &  \textbf{66.7} &  \textbf{77.8} &  \textbf{66.7} &  \textbf{55.6} &  \textbf{66.7} &  \textbf{77.8} &  \textbf{66.7} &  \textbf{66.7} &  \textbf{44.4} &          33.3  & 64.6 \\
\rowcolor{Gray}
Negation                      &   20 & \textbf{100.0} & \textbf{100.0} & \textbf{100.0} & \textbf{100.0} & \textbf{100.0} &  \textbf{95.0} & \textbf{100.0} & \textbf{100.0} &  \textbf{95.0} &          80.0  & \textbf{100.0} & 97.3 \\
\rowcolor{Gray}
Non-verbal agreement          &   61 &  \textbf{91.8} &  \textbf{88.5} &  \textbf{88.5} &  \textbf{86.9} &  \textbf{90.2} &  \textbf{83.6} &  \textbf{82.0} &  \textbf{88.5} &  \textbf{85.2} &          54.1  &          57.4  & 81.5 \\
Coreference                   &   20 &          80.0  &          70.0  &          75.0  &          80.0  &          75.0  &          70.0  &          70.0  &          75.0  &          70.0  &          70.0  &          60.0  & 72.3 \\
External possessor            &   21 & \textbf{100.0} &  \textbf{95.2} &  \textbf{90.5} &          85.7  &  \textbf{95.2} &          81.0  &          81.0  &  \textbf{90.5} &  \textbf{90.5} &          14.3  &          42.9  & 78.8 \\
Internal possessor            &   20 &  \textbf{95.0} & \textbf{100.0} & \textbf{100.0} &  \textbf{95.0} & \textbf{100.0} & \textbf{100.0} &  \textbf{95.0} & \textbf{100.0} &  \textbf{95.0} &          80.0  &          70.0  & 93.6 \\
\rowcolor{Gray}
\pagebreak
Punctuation                   &   60 &  \textbf{96.7} &  \textbf{98.3} &  \textbf{98.3} &          71.7  &          61.7  & \textbf{100.0} &  \textbf{98.3} &          70.0  &          28.3  &          68.3  &          55.0  & 77.0 \\
Comma                         &   20 & \textbf{100.0} & \textbf{100.0} & \textbf{100.0} &  \textbf{90.0} & \textbf{100.0} & \textbf{100.0} & \textbf{100.0} & \textbf{100.0} &          85.0  &          85.0  &          85.0  & 95.0 \\
Quotation marks               &   40 &  \textbf{95.0} &  \textbf{97.5} &  \textbf{97.5} &          62.5  &          42.5  & \textbf{100.0} &  \textbf{97.5} &          55.0  &           0.0  &          60.0  &          40.0  & 68.0 \\
\rowcolor{Gray}
Subordination                 &  180 &  \textbf{90.6} &  \textbf{88.3} &  \textbf{91.1} &  \textbf{91.1} &  \textbf{92.2} &  \textbf{88.9} &  \textbf{90.0} &  \textbf{90.6} &  \textbf{87.8} &          65.0  &          62.2  & 85.3 \\
Adverbial clause              &   20 &          90.0  &          90.0  &          90.0  &          90.0  &          95.0  &          90.0  &          90.0  &          85.0  &          90.0  &          85.0  &          80.0  & 88.6 \\
Cleft sentence                &   20 &  \textbf{95.0} &  \textbf{90.0} &  \textbf{90.0} &  \textbf{95.0} &  \textbf{95.0} &  \textbf{90.0} &  \textbf{95.0} &  \textbf{95.0} &  \textbf{80.0} &          55.0  &  \textbf{75.0} & 86.8 \\
Free relative clause          &   20 &  \textbf{90.0} &  \textbf{95.0} &  \textbf{95.0} &  \textbf{95.0} &  \textbf{90.0} &  \textbf{90.0} &  \textbf{85.0} &  \textbf{90.0} &  \textbf{85.0} &          65.0  &          60.0  & 85.5 \\
Indirect speech               &   20 &  \textbf{95.0} &          85.0  &  \textbf{90.0} &          85.0  & \textbf{100.0} &  \textbf{90.0} &  \textbf{90.0} &  \textbf{95.0} &          80.0  &          50.0  &          70.0  & 84.5 \\
Infinitive clause             &   20 & \textbf{100.0} & \textbf{100.0} & \textbf{100.0} &  \textbf{95.0} & \textbf{100.0} & \textbf{100.0} & \textbf{100.0} & \textbf{100.0} &  \textbf{95.0} &          80.0  &          65.0  & 94.1 \\
Object clause                 &   20 &  \textbf{95.0} &  \textbf{95.0} &  \textbf{95.0} & \textbf{100.0} & \textbf{100.0} &  \textbf{95.0} &  \textbf{95.0} & \textbf{100.0} &  \textbf{95.0} &          75.0  &          70.0  & 92.3 \\
Pseudo-cleft sentence         &   20 &  \textbf{70.0} &  \textbf{70.0} &  \textbf{70.0} &  \textbf{75.0} &  \textbf{80.0} &  \textbf{75.0} &  \textbf{70.0} &  \textbf{75.0} &  \textbf{75.0} &          40.0  &          35.0  & 66.8 \\
Relative clause               &   20 &  \textbf{85.0} &  \textbf{80.0} &  \textbf{95.0} &  \textbf{90.0} &  \textbf{85.0} &  \textbf{75.0} &  \textbf{90.0} &  \textbf{80.0} &  \textbf{90.0} &  \textbf{80.0} &          65.0  & 83.2 \\
Subject clause                &   20 &  \textbf{95.0} &  \textbf{90.0} &  \textbf{95.0} &  \textbf{95.0} &          85.0  &  \textbf{95.0} &  \textbf{95.0} &  \textbf{95.0} & \textbf{100.0} &          55.0  &          40.0  & 85.5 \\
\rowcolor{Gray}
Verb tense/aspect/mood        & 4447 &  \textbf{84.6} &  \textbf{85.3} &          80.3  &          75.9  &          79.6  &          77.5  &          75.1  &          79.3  &          73.6  &          50.5  &          52.1  & 74.0 \\
Conditional                   &   19 & \textbf{100.0} & \textbf{100.0} & \textbf{100.0} & \textbf{100.0} &  \textbf{89.5} &  \textbf{89.5} &  \textbf{94.7} &  \textbf{94.7} &          84.2  &          73.7  &          84.2  & 91.9 \\
Ditransitive - future I       &   36 & \textbf{100.0} & \textbf{100.0} & \textbf{100.0} & \textbf{100.0} & \textbf{100.0} & \textbf{100.0} & \textbf{100.0} & \textbf{100.0} &          91.7  &          75.0  &          66.7  & 93.9 \\
Ditransitive - future I subjunctive II &   36 & \textbf{100.0} & \textbf{100.0} & \textbf{100.0} & \textbf{100.0} & \textbf{100.0} & \textbf{100.0} & \textbf{100.0} & \textbf{100.0} & \textbf{100.0} &          75.0  &          55.6  & 93.7 \\
Ditransitive - future II      &   36 & \textbf{100.0} & \textbf{100.0} & \textbf{100.0} & \textbf{100.0} &          86.1  & \textbf{100.0} &  \textbf{97.2} & \textbf{100.0} & \textbf{100.0} &          69.4  &          91.7  & 94.9 \\
Ditransitive - future II subjunctive II &   36 & \textbf{100.0} & \textbf{100.0} & \textbf{100.0} & \textbf{100.0} & \textbf{100.0} &  \textbf{97.2} & \textbf{100.0} & \textbf{100.0} & \textbf{100.0} &          69.4  &  \textbf{94.4} & 96.5 \\
Ditransitive - perfect        &   36 & \textbf{100.0} & \textbf{100.0} & \textbf{100.0} & \textbf{100.0} & \textbf{100.0} & \textbf{100.0} & \textbf{100.0} & \textbf{100.0} &  \textbf{97.2} &          83.3  &          86.1  & 97.0 \\
Ditransitive - pluperfect     &   36 &  \textbf{94.4} &  \textbf{94.4} &          66.7  &          86.1  &          80.6  & \textbf{100.0} &          22.2  &          63.9  &          33.3  &          44.4  &          50.0  & 66.9 \\
Ditransitive - pluperfect subjunctive II &   36 & \textbf{100.0} &          91.7  & \textbf{100.0} & \textbf{100.0} & \textbf{100.0} & \textbf{100.0} & \textbf{100.0} & \textbf{100.0} &  \textbf{97.2} &          33.3  &          63.9  & 89.6 \\
Ditransitive - present        &   36 &  \textbf{97.2} &  \textbf{97.2} & \textbf{100.0} & \textbf{100.0} & \textbf{100.0} &  \textbf{97.2} &          72.2  &  \textbf{97.2} &          83.3  &          75.0  &          63.9  & 89.4 \\
Ditransitive - preterite      &   36 &  \textbf{88.9} &  \textbf{86.1} &          77.8  &  \textbf{97.2} &          77.8  &          77.8  &          75.0  &  \textbf{94.4} &          83.3  &          50.0  &          50.0  & 78.0 \\
Ditransitive - preterite subjunctive II &   36 &  \textbf{75.0} &  \textbf{75.0} &  \textbf{69.4} &  \textbf{72.2} &  \textbf{63.9} &  \textbf{66.7} &  \textbf{66.7} &  \textbf{72.2} &  \textbf{77.8} &          36.1  &          44.4  & 65.4 \\
Imperative                    &   20 &  \textbf{85.0} &  \textbf{80.0} &  \textbf{80.0} &  \textbf{95.0} &  \textbf{95.0} &  \textbf{75.0} &  \textbf{80.0} &  \textbf{90.0} &  \textbf{85.0} &          30.0  &          35.0  & 75.5 \\
Intransitive - future I       &   36 &  \textbf{97.2} &  \textbf{97.2} &  \textbf{97.2} &  \textbf{97.2} & \textbf{100.0} &  \textbf{97.2} & \textbf{100.0} &  \textbf{97.2} &  \textbf{97.2} &          86.1  &          66.7  & 93.9 \\
Intransitive - future I subjunctive II &   36 & \textbf{100.0} & \textbf{100.0} & \textbf{100.0} &          86.1  & \textbf{100.0} & \textbf{100.0} & \textbf{100.0} & \textbf{100.0} & \textbf{100.0} &          80.6  &          69.4  & 94.2 \\
Intransitive - future II      &   42 &          88.1  &          90.5  &  \textbf{95.2} &  \textbf{95.2} &          88.1  &          92.9  & \textbf{100.0} &  \textbf{97.6} &          92.9  &          61.9  &          35.7  & 85.3 \\
Intransitive - future II subjunctive II &   36 & \textbf{100.0} & \textbf{100.0} & \textbf{100.0} & \textbf{100.0} &          61.1  & \textbf{100.0} & \textbf{100.0} & \textbf{100.0} & \textbf{100.0} &          44.4  &          36.1  & 85.6 \\
Intransitive - perfect        &   84 & \textbf{100.0} & \textbf{100.0} & \textbf{100.0} & \textbf{100.0} &  \textbf{98.8} & \textbf{100.0} & \textbf{100.0} & \textbf{100.0} &  \textbf{97.6} &          63.1  &          54.8  & 92.2 \\
Intransitive - pluperfect     &   36 &  \textbf{83.3} &  \textbf{83.3} &  \textbf{66.7} &          33.3  &  \textbf{83.3} &  \textbf{75.0} &          50.0  &  \textbf{77.8} &          61.1  &          38.9  &          19.4  & 61.1 \\
Intransitive - pluperfect subjunctive II &   36 & \textbf{100.0} & \textbf{100.0} &  \textbf{97.2} &  \textbf{94.4} & \textbf{100.0} & \textbf{100.0} &  \textbf{94.4} &  \textbf{97.2} & \textbf{100.0} &          25.0  &          19.4  & 84.3 \\
Intransitive - present        &   36 & \textbf{100.0} & \textbf{100.0} & \textbf{100.0} &  \textbf{97.2} & \textbf{100.0} & \textbf{100.0} & \textbf{100.0} & \textbf{100.0} &  \textbf{97.2} &          69.4  &          52.8  & 92.4 \\
Intransitive - preterite      &   66 &  \textbf{93.9} &  \textbf{97.0} &          86.4  &  \textbf{92.4} &  \textbf{92.4} &  \textbf{95.5} &          84.8  &  \textbf{97.0} &  \textbf{95.5} &          47.0  &          19.7  & 82.0 \\
Intransitive - preterite subjunctive II &   36 &          63.9  &  \textbf{75.0} &          63.9  &  \textbf{77.8} &  \textbf{80.6} &          61.1  &  \textbf{88.9} &  \textbf{75.0} &          66.7  &          22.2  &          11.1  & 62.4 \\
Modal - future I              &  180 &          79.4  &  \textbf{90.0} &          73.9  &          78.9  &          79.4  &          73.9  &          72.2  &          75.6  &          75.6  &          62.2  &          59.4  & 74.6 \\
Modal - future I subjunctive II &  180 &          77.2  &  \textbf{86.7} &          76.1  &          76.7  &          74.4  &          72.8  &          65.0  &          66.1  &          70.6  &          57.2  &          64.4  & 71.6 \\
Modal - perfect               &  180 &  \textbf{90.0} &  \textbf{83.9} &  \textbf{85.0} &          73.9  &          70.6  &          65.0  &          82.8  &          78.3  &          60.0  &           1.7  &          63.3  & 68.6 \\
Modal - pluperfect            &  180 &  \textbf{50.6} &  \textbf{40.6} &          37.2  &          22.8  &          32.8  &          15.6  &          28.3  &          16.7  &           2.2  &           0.0  &           8.9  & 23.2 \\
Modal - pluperfect subjunctive II &  180 &  \textbf{65.6} &  \textbf{60.6} &  \textbf{60.0} &          43.9  &  \textbf{59.4} &  \textbf{56.7} &  \textbf{60.0} &          54.4  &  \textbf{55.0} &          37.2  &          32.8  & 53.2 \\
Modal - present               &  180 &  \textbf{97.8} &  \textbf{96.1} &          93.3  &          80.0  &  \textbf{94.4} &          91.7  &          69.4  &  \textbf{95.0} &          92.2  &          81.7  &          69.4  & 87.4 \\
Modal - preterite             &  180 &  \textbf{98.3} &  \textbf{99.4} &          94.4  &          96.7  &  \textbf{98.9} &          96.7  &          93.9  &  \textbf{98.9} &  \textbf{97.8} &          80.6  &          72.8  & 93.5 \\
Modal - preterite subjunctive II &  180 &  \textbf{73.9} &  \textbf{78.9} &  \textbf{75.0} &          60.0  &  \textbf{75.6} &  \textbf{78.9} &  \textbf{77.8} &  \textbf{77.2} &          61.7  &          57.8  &          52.2  & 69.9 \\
Modal negated - future I      &  180 &          80.0  &  \textbf{92.8} &          75.6  &          78.3  &          79.4  &          76.1  &          75.6  &          79.4  &          77.2  &          66.7  &          67.2  & 77.1 \\
Modal negated - future I subjunctive II &  180 &  \textbf{78.3} &  \textbf{85.6} &          76.1  &          75.6  &          76.1  &          70.6  &          69.4  &          77.2  &          69.4  &          57.8  &          63.3  & 72.7 \\
Modal negated - perfect       &  171 &  \textbf{95.9} &  \textbf{93.6} &  \textbf{94.2} &          70.2  &          73.7  &          70.2  &          88.3  &          80.7  &          70.2  &           9.4  &          67.8  & 74.0 \\
Modal negated - pluperfect    &  169 &  \textbf{36.1} &  \textbf{22.5} &          18.3  &  \textbf{32.0} &  \textbf{32.5} &          14.8  &          12.4  &           4.1  &           0.0  &   \textbf{0.6} &           7.1  & 16.4 \\
Modal negated - pluperfect subjunctive II &  179 &  \textbf{71.5} &  \textbf{70.9} &  \textbf{73.7} &          40.8  &          52.5  &          57.0  &          53.6  &          63.1  &          54.2  &          31.8  &          34.6  & 54.9 \\
Modal negated - present       &  169 & \textbf{100.0} & \textbf{100.0} &  \textbf{99.4} &          84.6  &          92.3  &  \textbf{98.8} &          76.3  &  \textbf{99.4} &          84.0  &          81.7  &          82.8  & 90.9 \\
Modal negated - preterite     &  179 &  \textbf{99.4} & \textbf{100.0} &          96.6  &          91.1  &  \textbf{98.9} &  \textbf{99.4} &          92.7  & \textbf{100.0} &          88.3  &          74.3  &          76.0  & 92.4 \\
Modal negated - preterite subjunctive II &  174 &  \textbf{74.7} &  \textbf{81.0} &          70.1  &          62.6  &  \textbf{79.9} &  \textbf{81.6} &  \textbf{75.3} &          70.1  &  \textbf{75.9} &          54.6  &          50.0  & 70.5 \\
Progressive                   &   19 &  \textbf{84.2} &  \textbf{73.7} &  \textbf{63.2} &  \textbf{78.9} &  \textbf{73.7} &  \textbf{84.2} &  \textbf{78.9} &  \textbf{84.2} &  \textbf{63.2} &          15.8  &          36.8  & 67.0 \\
Reflexive - future I          &   36 & \textbf{100.0} & \textbf{100.0} &          91.7  &          88.9  &          88.9  &  \textbf{94.4} &          80.6  &  \textbf{94.4} &  \textbf{97.2} &          69.4  &          33.3  & 85.4 \\
Reflexive - future I subjunctive II &   36 &          83.3  &  \textbf{86.1} &          83.3  &  \textbf{91.7} &          80.6  &          83.3  &          69.4  &  \textbf{88.9} &  \textbf{97.2} &          55.6  &          30.6  & 77.3 \\
Reflexive - future II         &   36 &  \textbf{94.4} & \textbf{100.0} &          88.9  &          91.7  &          91.7  &          86.1  &          77.8  &  \textbf{94.4} &          69.4  &          41.7  &          27.8  & 78.5 \\
Reflexive - future II subjunctive II &   36 &  \textbf{83.3} &  \textbf{88.9} &  \textbf{83.3} &  \textbf{91.7} &          61.1  &  \textbf{83.3} &          75.0  &  \textbf{86.1} &  \textbf{86.1} &          30.6  &          25.0  & 72.2 \\
Reflexive - perfect           &   36 & \textbf{100.0} & \textbf{100.0} &  \textbf{94.4} &  \textbf{94.4} &  \textbf{94.4} &          91.7  &          86.1  &          91.7  &          91.7  &          41.7  &          33.3  & 83.6 \\
Reflexive - pluperfect        &   36 &          91.7  & \textbf{100.0} &          83.3  &          86.1  &          91.7  &          80.6  &          80.6  &  \textbf{97.2} &          91.7  &          30.6  &          19.4  & 77.5 \\
Reflexive - pluperfect subjunctive II &   36 &  \textbf{86.1} &  \textbf{83.3} &  \textbf{80.6} &  \textbf{91.7} &  \textbf{88.9} &  \textbf{80.6} &          75.0  &  \textbf{80.6} &  \textbf{88.9} &          36.1  &          22.2  & 74.0 \\
Reflexive - present           &   36 &  \textbf{97.2} & \textbf{100.0} &          91.7  &          91.7  &          91.7  &          86.1  &          69.4  &  \textbf{94.4} &          88.9  &          33.3  &          13.9  & 78.0 \\
Reflexive - preterite         &   36 &  \textbf{94.4} &  \textbf{86.1} &          69.4  &  \textbf{91.7} &  \textbf{86.1} &          69.4  &          75.0  &  \textbf{88.9} &          75.0  &          13.9  &          16.7  & 69.7 \\
Reflexive - preterite subjunctive II &   36 &  \textbf{83.3} &  \textbf{75.0} &          61.1  &  \textbf{77.8} &  \textbf{80.6} &          61.1  &  \textbf{69.4} &  \textbf{75.0} &  \textbf{66.7} &          22.2  &          19.4  & 62.9 \\
Transitive - future I         &   42 &  \textbf{97.6} &  \textbf{97.6} &  \textbf{97.6} &  \textbf{97.6} &  \textbf{97.6} &  \textbf{97.6} &  \textbf{97.6} &  \textbf{97.6} &  \textbf{95.2} &  \textbf{90.5} &          81.0  & 95.2 \\
Transitive - future I subjunctive II &   36 & \textbf{100.0} & \textbf{100.0} & \textbf{100.0} & \textbf{100.0} & \textbf{100.0} & \textbf{100.0} & \textbf{100.0} & \textbf{100.0} & \textbf{100.0} &  \textbf{97.2} &          77.8  & 97.7 \\
Transitive - future II        &   36 & \textbf{100.0} & \textbf{100.0} & \textbf{100.0} & \textbf{100.0} & \textbf{100.0} & \textbf{100.0} & \textbf{100.0} & \textbf{100.0} & \textbf{100.0} &          91.7  &          63.9  & 96.0 \\
Transitive - future II subjunctive II &   36 & \textbf{100.0} & \textbf{100.0} & \textbf{100.0} & \textbf{100.0} & \textbf{100.0} & \textbf{100.0} & \textbf{100.0} & \textbf{100.0} & \textbf{100.0} &  \textbf{97.2} &          63.9  & 96.5 \\
Transitive - perfect          &   42 & \textbf{100.0} & \textbf{100.0} & \textbf{100.0} & \textbf{100.0} &  \textbf{97.6} & \textbf{100.0} & \textbf{100.0} & \textbf{100.0} &  \textbf{95.2} &          85.7  &          66.7  & 95.0 \\
Transitive - pluperfect       &   36 & \textbf{100.0} & \textbf{100.0} &          77.8  &          80.6  & \textbf{100.0} & \textbf{100.0} &          75.0  &  \textbf{94.4} &          80.6  &          66.7  &          36.1  & 82.8 \\
Transitive - pluperfect subjunctive II &   36 & \textbf{100.0} &          91.7  & \textbf{100.0} & \textbf{100.0} & \textbf{100.0} & \textbf{100.0} & \textbf{100.0} & \textbf{100.0} & \textbf{100.0} &          30.6  &          44.4  & 87.9 \\
Transitive - present          &   48 & \textbf{100.0} & \textbf{100.0} & \textbf{100.0} & \textbf{100.0} & \textbf{100.0} & \textbf{100.0} &  \textbf{95.8} & \textbf{100.0} &  \textbf{97.9} &          75.0  &          75.0  & 94.9 \\
Transitive - preterite        &   36 &  \textbf{97.2} &  \textbf{97.2} &          88.9  &          91.7  &  \textbf{97.2} &          88.9  &          91.7  & \textbf{100.0} &  \textbf{97.2} &          66.7  &          55.6  & 88.4 \\
Transitive - preterite subjunctive II &   36 &  \textbf{66.7} &  \textbf{75.0} &  \textbf{63.9} &  \textbf{69.4} &          61.1  &          58.3  &  \textbf{80.6} &  \textbf{83.3} &  \textbf{69.4} &          25.0  &          50.0  & 63.9 \\
\rowcolor{Gray}
Verb valency                  &   87 &  \textbf{79.3} &  \textbf{81.6} &  \textbf{77.0} &  \textbf{81.6} &  \textbf{77.0} &  \textbf{77.0} &  \textbf{71.3} &  \textbf{80.5} &          64.4  &          44.8  &          51.7  & 71.5 \\
Case government               &   28 &  \textbf{92.9} &  \textbf{96.4} &  \textbf{85.7} &  \textbf{92.9} &  \textbf{82.1} &  \textbf{89.3} &          78.6  &  \textbf{92.9} &  \textbf{82.1} &          42.9  &          57.1  & 81.2 \\
Mediopassive voice            &   20 &  \textbf{85.0} &  \textbf{90.0} &  \textbf{85.0} &  \textbf{95.0} &  \textbf{90.0} &  \textbf{80.0} &          70.0  &  \textbf{90.0} &          55.0  &          50.0  &          50.0  & 76.4 \\
Passive voice                 &   19 & \textbf{100.0} & \textbf{100.0} & \textbf{100.0} & \textbf{100.0} & \textbf{100.0} & \textbf{100.0} & \textbf{100.0} & \textbf{100.0} &  \textbf{89.5} &          78.9  &          73.7  & 94.7 \\
Resultative predicates        &   20 &          35.0  &          35.0  &          35.0  &          35.0  &          35.0  &          35.0  &          35.0  &          35.0  &          25.0  &          10.0  &          25.0  & 30.9 \\
\midrule
micro-average                 & 5514 &  \textbf{85.3} &  \textbf{85.4} &          81.2  &          77.7  &          80.6  &          78.7  &          76.5  &          79.1  &          73.6  &          51.3  &          52.4  & 74.7 \\
phenomenon macro-average     & 5514 &  \textbf{89.1} &          88.0  &          85.2  &          85.1  &          85.5  &          83.9  &          82.1  &          83.7  &          78.5  &          53.7  &          51.8  & 78.8 \\
category macro-average        & 5514 &  \textbf{88.1} &          86.8  &          85.3  &          84.6  &          84.3  &          83.6  &          82.7  &          80.0  &          74.1  &          54.9  &          54.4  & 78.1 \\
BLEU & & 43.8 & 43.5 & 42.3 & 41.9 & 41.4 & 40.4 & 39.6 & 43.2 & 35.4 & 31.5 & 32.1 & 35.4 \\
\bottomrule

\caption{Accuracies (\%) of successful translations for 11 systems regarding all phenomena, organized in categories.
      Boldface indicates the best scoring system in each row, including all systems which are not significantly inferior than the best scoring system. Grey rows average the accuracies of the phenomena per category.}
\label{tab:phenomena}
\end{longtable}

}

\end{landscape}

\end{document}